\documentclass[letterpaper, 10 pt, conference]{ieeeconf}  

\IEEEoverridecommandlockouts                              
\usepackage{amsmath,amssymb,amsfonts,booktabs,multirow,bm,balance,bbm,siunitx,hyperref,graphicx}
\usepackage[caption=false, font=footnotesize]{subfig}
\usepackage{commath}
\usepackage{graphicx}
\usepackage{float}
\usepackage{balance}
\usepackage{xcolor}

\usepackage[ruled,vlined]{algorithm2e}
\usepackage[font=footnotesize,skip=1pt]{caption}
\graphicspath{ {./figures)/} }

\usepackage{xcolor}

\title{\LARGE \bf
A collision-resilient aerial vehicle with icosahedron tensegrity structure}

\author{Jiaming Zha, Xiangyu Wu, Joseph Kroeger, Natalia Perez and Mark W. Mueller
\thanks{The authors are with the HiPeRLab, University of California, Berkeley, CA 94720, USA. {\tt\small \{jiaming\_zha,wuxiangyu,jrkroeger,{\break}nataliaperez,mwm\}@berkeley.edu} }
}

\begin{document}
\maketitle
\pagestyle{empty}
\begin{abstract} 
Aerial vehicles with collision resilience can operate with more confidence in environments with obstacles that are hard to detect and avoid.
This paper presents the methodology used to design a collision resilient aerial vehicle with icosahedron tensegrity structure. 
A simplified stress analysis of the tensegrity frame under impact forces is performed to guide the selection of its components. 
In addition, an autonomous controller is presented to reorient the vehicle from an arbitrary orientation on the ground to help it take off. 
Experiments show that the vehicle can successfully reorient itself after landing upside-down and can survive collisions with speed up to 6.5m/s.
\end{abstract}

\section{Introduction}
Unmanned Aerial Vehicles (UAVs) have been used for search and rescue missions in environments that are dangerous (e.g.  \cite{waharte2010supporting,scherer2015autonomous}). These hazardous environments, such as a tunnel after an earthquake, often challenge UAVs with complex obstacles that are hard to detect and difficult to avoid. Collisions with these obstacles can disrupt missions and damage vehicles. 

To address this problem, many collision resilient designs are proposed to help aerial vehicles operate safely in a wide range of environments. Most of these designs fall into one of three major categories: 
\begin{enumerate}
  \item Adding additional propeller guards \cite{salaan2019development,snap1}.
  \item Covering the whole vehicle with an external impact resilient shell \cite{briod2014collision,kornatowski2017origami,elios2}.
  \item Using novel materials that decrease the stiffness of the vehicle during collisions \cite{mintchev2017insect,jang2019design}. 
\end{enumerate}
 
Resilient designs of the first category protect propellers, a key yet vulnerable part of aerial vehicles, from damage in collisions. For example, spherical guards that can passively rotate about vehicle propellers are proposed in \cite{salaan2019development}. The design prevents the vehicle from bending moment during collisions and thus increases its surviving rate. 

Meanwhile, designs with external shells provide an all-around protection. For instance, a collision resilient flying robot protected by a spherical shell with an inner gimbal system is proposed in \cite{briod2014collision}. The design decouples its outer protective shell from its inner propulsion system. As a result, the inner frame can remain stable during collisions. On the other hand, a cargo drone with an origami-inspired external protective shell is demonstrated in \cite{kornatowski2017origami}. The vehicle can be folded to reduce its storage volume. 

\begin{figure}[b]
    \centering
    \includegraphics[width=0.7\linewidth]{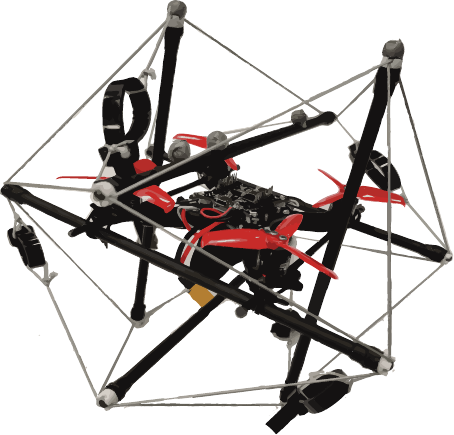}
    \caption{The icosahedron structured aerial vehicle. The total mass is 252g and the length of the rod in the tensegrity is 20cm.}
    \label{VehiclePic}
\end{figure}

Collision resilient designs of the last category utilize novel materials. A quadcopter frame with dual stiffness materials is proposed in \cite{mintchev2017insect}. The frame stays rigid during normal flights, and becomes soft to absorb energy once it collides. Similarly, a flexible propeller blade that bends when colliding with obstacles is presented in \cite{jang2019design}. 

Unfortunately, most of the existing collision resilient aerial vehicles are not designed to survive collisions at high-speed, which robotic first responders may wish to operate at. For example, when a UAV is passing a corridor on fire, it needs to rush through the corridor so as to avoid long time exposure to the heat. A fast flight speed is also desired when the rescue mission is time-critical. 

In this work we show that vehicles constructed with icosahedron tensegrity (tensile-integrity) structures have the potential to survive collisions during fast flights. A tensegrity structure is constructed with rigid bodies suspended in a tension network. The unique static and dynamic features of tensegrity structures have been studied by many (e.g. \cite{cheong2015nonminimal,murakami2001static}). Due to these features, an icosahedron tensegrity structure can distribute load among its members and does not experience bending moment in collisions. As a result, the structure can provide high impact resilience with a light weight, and it has consequently been proposed as a potential candidate for planetary landers \cite{rimoli2016impact} and exploratory rovers \cite{sunspiral2013tensegrity,kim2016hopping}. 

The contribution of this work is as follows. We present the methodology used to design a collision resilient aerial vehicle that takes advantage of the high impact-tolerance of tensegrity structures. We introduce an analysis tool that approximates the stress in the tensegrity frame during collisions and use the tool to guide the selection of tensegrity components. In addition, we propose a controller that can reorient the vehicle from an arbitrary orientation on the ground to help it take off. The proposed controller exploits the 20-faced geometry of the icosahedron tensegrity and divides the task into simple rotations that are easy to analyze and implement.

With the proposed methodology, we achieve a new collision resilient vehicle that can safely operate at high speed despite the presence of obstacles and can resume flight after collisions. The vehicle, shown in Fig. \ref{VehiclePic}, is protected by an orthogonal icosahedron \cite{jessen1967orthogonal} tensegrity shell made with carbon fiber rods and high tensile braided fishing lines, which provide collision resilience with a light weight. Experiments show that the vehicle can survive collisions with speed up to 6.5m/s. To the best of our knowledge, no other aerial vehicles with impact resilient designs in literature has reported to survive collisions with a higher speed.

\section{Design of the tensegrity structure}
In this section, we introduce the design objectives of the tensegrity structure, the stress analysis used to guide the design and the criteria for selecting tensegrity components.

\subsection{Design objectives}
The goal of the tensegrity structure is to protect the quadcopter inside, which has to survive collisions and continue its operation. As a result, the deformation of the tensegrity structure should be small during the collision so that vulnerable parts like sensors and propellers are not exposed to obstacles.
Moreover, the tensegrity structure should also be lightweight, so that the additional mass of the structure has limited influence on the vehicle's flight performance and payload capacity.

\subsection{Simplified stress analysis of icosahedron tensegrity}

The design of the icosahedron tensegrity structure is guided by the stress its components experience during the collisions.
To simplify the analysis, we consider only the moment when the vehicle is experiencing the largest impact force from the obstacle. 
To further simplify the problem, we additionally assume that: 1) the vehicle's weight is small compared to the impact force and can be neglected; 2) the deformation of the tensegrity structure is small so that the geometry of the tensegrity structure remains unchanged. The second assumption is a close approximation, because stiff strings are used in the design to prevent large deformations that may expose vulnerable parts like propellers to collisions.

\begin{figure}
    \centering
    \includegraphics[width=\linewidth]{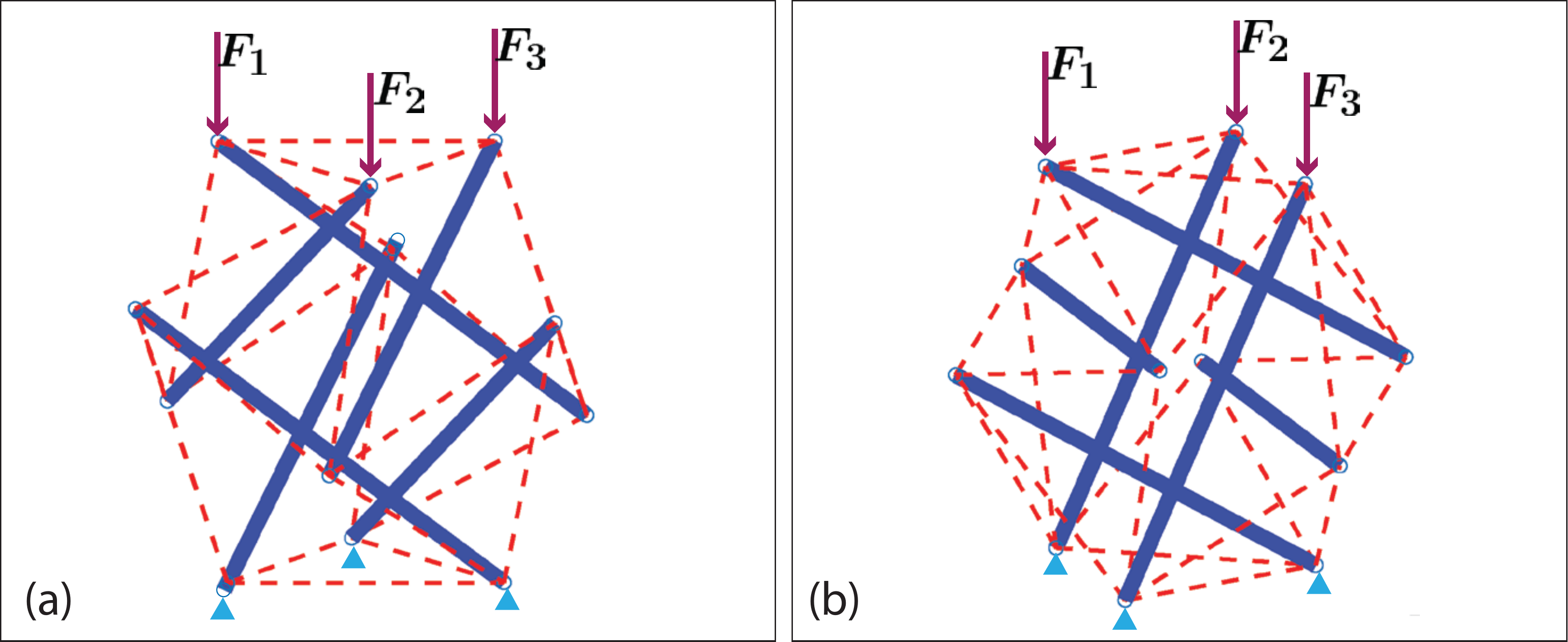}
    \caption{Setup of the simplified stress analysis problem: the bottom of the tensegrity is connected to the ground, and forces are exerted to tensegrity's top nodes. Two configurations are considered: (a) Forces act on a face with three string-pieces. (b) Forces act on a face with two string-pieces.}
    \label{StressAnalysis}
\end{figure}

Forces in the tensegrity structure can thus be approximately solved from the following problem. Shown in Fig. \ref{StressAnalysis}, an icosahedron tensegrity is fixed on the ground with three connections. Vertical forces with a sum of $F_{max}$, the maximum impact force in a collision, is applied to the top nodes of the vehicle. 
Due to symmetry, only two configurations need to be considered: forces act on a face with three string-pieces or on a face with two string-pieces. 
Moreover, for each configuration, we consider following scenarios: the total impact force is evenly distributed among 1, 2 or 3 nodes.

The equilibrium equations of the system can then be formulated. The icosahedron tensegrity has 12 nodes. Denote the nodes as $\boldsymbol{n}_i$, where $i = 1,...,12$. These nodes are connected by 6 rods and 20 string-pieces. We denote the connectivity of the nodes with functions $N_r$ and $N_s$:
\begin{flalign}
 \label{eq1}
 & N_r(i,j)=   \begin{cases}
                                   k \mbox{, if rod $k$ connects
                                   $\boldsymbol{n}_i$ and $\boldsymbol{n}_j$} \\
                                   0 \mbox{, otherwise} 
  \end{cases} 
  \\
 \label{eq2}
 & N_s(i,j)=   \begin{cases}
                                   l \mbox{, if string-piece $l$ connects $\boldsymbol{n}_i$ and $\boldsymbol{n}_j$} \\
                                   0 \mbox{, otherwise} \\
  \end{cases}
\end{flalign}

Denote $T_{l}$ as the magnitude of the tensile force in string-piece $l$, and $C_{k}$ as the magnitude of the compression force in rod $k$. Use $\boldsymbol{U}^B_{i} \in \mathbb{R}^{3}$ to represent the external force on $\boldsymbol{n}_i$ in the vehicle body frame. Define  $\boldsymbol{s}^B_{i,j}$ as the unit vector along the string-piece from $\boldsymbol{n}_i$ to $\boldsymbol{n}_j$, whereas $\boldsymbol{r}^B_{i,j}$ as the unit vector along the rod from $\boldsymbol{n}_i$ to $\boldsymbol{n}_j$. Both vectors are in the vehicle body frame.

At equilibrium, the sum of forces acting on each node is zero:
\begin{align}
\sum \boldsymbol{F_i} = 0
 \label{Summation of force}
\end{align} 
This equation comes from Newton's law with the assumption that each node has zero acceleration. A more rigorous derivation of tensegrity equilibrium can be found in \cite{goyal2018dynamics}. Expand the equation, we have for each node $i$:
\begin{align}
\sum_{\{ j|N_r(i,j)\ne 0 \}}-C_k\boldsymbol{r}^B_{i,j}+\sum_{\{ j|N_s(i,j)\ne 0 \}} T_l\boldsymbol{s}^B_{i,j}+\boldsymbol{U}^B_{i} = 0
 \label{ForceEq}
\end{align} 
where the indices of string-pieces, $l$ and the indices of rods, $k$, come from connectivity functions: $l = N_s(i,j)$ and $k = N_r(i,j)$.

Due to the existence of self-stress in the icosahedron tensegrity, the structure is statically indeterminate \cite{pellegrino1986matrix}. With the assumption that the members of the tensegrity structure follow Hooke's law when the deformation is small, we apply the principle of minimum energy and solve for a minimum energy configuration solution under equilibrium constraints, with a method similar to the one presented in \cite{monforton1980analysis}:

\begin{align}
\begin{aligned}
\min_{T,C,\boldsymbol{U}^B} & \sum_{k} \frac{1}{2} K_{r}(\epsilon_{k}L_r)^2 +  \sum_{l} \frac{1}{2} K_{s}(\epsilon_{l}L_s)^2\\
\\
\textrm{s.t.  } & \text{Equilibrium condition: Eq. \eqref{ForceEq} }
\\
&  \epsilon_{k} = \frac{C_k}{E_sA_s}\\
&  \epsilon_{l} = \frac{T_l}{E_rA_r}\geq 0\\
\label{optimization}
\end{aligned}
\end{align}
Constants $K$, $E$, $A$, $L$ denote Hooke's constant, Young's modulus, cross sectional area and length  respectively. Subscripts $s$ and $r$ represents string-pieces and rods. $\epsilon_k$ and $\epsilon_l$ are the strain of rod $k$ and string-piece $l$. Notice that the strain of the string-pieces is constrained to be positive as the string-pieces cannot be compressed. 

Finally, compare the results of all possible configurations and external force scenarios. Define the maximum compression force in rods as $C_{max}$ and maximum tensile force in string-pieces as $T_{max}$.

\subsection{Selection of tensegrity components}
Tensegrity components (rods and strings) in the structure should survive the maximum tensile and compression force found in the previous subsection. In other word, the stress in the string should be smaller than its yielding strength $\sigma_{ys}$ with a factor of safety $\eta_s$: 
\begin{align}
\eta_s \frac{T_{max}}{A_s}  < \sigma_{ys}
\label{StringTensionLimit}
\end{align}
Meanwhile, the maximum stress in the rod should be smaller than both its yielding strength $\sigma_{yr}$ and its critical buckling strength $\sigma_{br}$ with factors of safety $\eta_{r1}$ and $\eta_{r2}$. 
\begin{align}
\eta_{r1} \frac{C_{max}}{A_r}  < \sigma_{yr},  \eta_{r2} \frac{C_{max}}{A_r}  < \sigma_{br}
\label{RodCompressionLimit}
\end{align}
The critical buckling strength of the rod can be approximated with Euler's buckling theory: 
\begin{align}
\sigma_{br} = \frac{\pi^2E_rI_r}{A_r(L_r)^2}
\label{RodBucklingLimit}
\end{align}
Here $I_r$ is the second moment of area of the rod.

\section{Vehicle controllers and state estimators}

\subsection{Sensing and state estimation}
The vehicle's onboard sensing system is based on an Inertial Measurement Unit (IMU) that provides information about the vehicle's acceleration and angular velocities. When flying in the lab space, the vehicle is visible to a motion capture system, allowing for off-board estimation of the vehicle’s full 6 degrees of freedom position and orientation states. The onboard rate gyroscope is also used in flight for attitude estimation and angular velocity feedback.

During the reorientation process, however, the vehicle is assumed to lose access to the motion capture system and thus solely relies on its IMU for state estimation. It estimates its attitude from accelerometer and rate gyroscope measurements with a complementary filter.

\subsection{Flight controller}

\begin{figure}
    \centering
    \includegraphics[width=\linewidth]{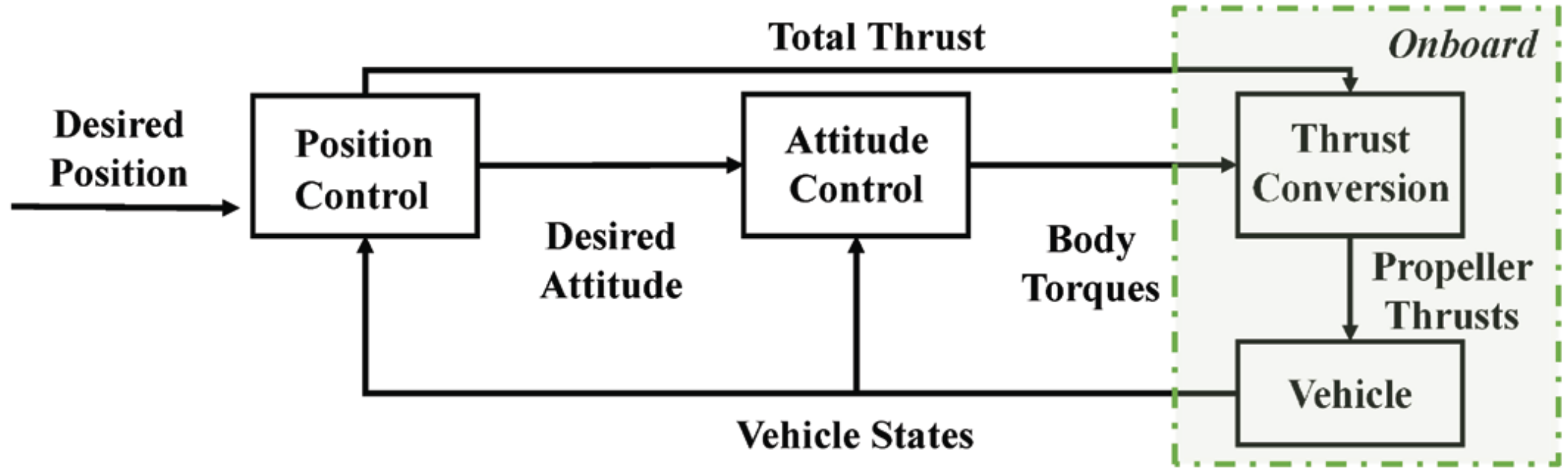}
    \caption{Control Architecture of the Vehicle.}
    \label{ControlDiagram}
\end{figure}

The overall control strategy of the vehicle is presented in Fig. \ref{ControlDiagram}. It features a cascaded control structure. A position controller outputs desired total thrust and thrust direction, whereas an inner attitude controller computes desired torques. Finally, a mixer converts the total thrust and body torque commands to per-propeller thrust commands. This cascaded structure can be decoupled into two separate parts: an offboard controller for position and attitude control, and an onboard part implementing thrust conversion.  

The position controller makes position error act as a second order system with damping ratio ${\zeta}_p$ and natural frequency ${\omega}_p$:
\begin{align}
\ddot{\boldsymbol d}^E_{d} = 2\zeta_p \omega_p (\dot{\boldsymbol d}^E_{d} - \dot{\boldsymbol d}^E) + \omega_p^2(\boldsymbol d_{d}^E - \boldsymbol d^E) 
\end{align}
wherein $\boldsymbol{d}^E$ is the position of the vehicle, $\ddot{\boldsymbol d}^E_{d}$ is desired acceleration, $\dot{\boldsymbol d}^E_{d}$ is desired velocity and ${\boldsymbol d}^E_{d}$ is desired position. All vectors are in the Earth frame. Thus, we can find desired total thrust and its direction as: 

\begin{equation}\label{eq:posSplit}
f_d = m||\ddot{\boldsymbol d}^E_{d}||_2, \qquad
\boldsymbol z_{B,d} = \frac{\ddot{\boldsymbol d}^E_{d}}{||\ddot{\boldsymbol d}^E_{d}||_2}
\end{equation}

A desired angular velocity $\boldsymbol{ \omega}_{d}^B$ is computed as proportional to the orientation error and then an desired angular acceleration is computed as follows:
\begin{equation}
\dot{\boldsymbol{ \omega}}_{d}^B =  \frac{1}{\tau}(\boldsymbol{\omega}_{d}^B - \boldsymbol{\omega}^B)
\end{equation}
where $\boldsymbol{\omega}^B$ is the angular velocity measured by rate gyroscope and $\tau$ is the desired time constant. All vectors are in vehicle's body frame.
Thus, the desired torque follows as: 
\begin{equation}
\boldsymbol{\tau}^B_d = \boldsymbol{J}\dot{\boldsymbol{ \omega}}_{d}^B
\end{equation}
Where $\boldsymbol{J}$ is the moment of inertia of the vehicle. 

Finally, the thrust converter computes the desired thrust force for each propeller as follows:
\begin{equation}\label{eq:mixer}
f_{P_i} = \frac{1}{4}\bigg(\begin{bmatrix}
r_{i,y}^{-1} & -r_{i,x}^{-1} & \kappa^{-1}
\end{bmatrix} \boldsymbol \tau^B_d + f_{d}\bigg)
\end{equation}
where $\boldsymbol{r_i}$ is a vector pointing from center of mass to propeller $i$ and $r_{i,x}$ and $r_{i,y}$ are the components of $\boldsymbol r_i$ along the x-axis and y-axis of the body frame. $\kappa$ is the propeller torque constant.

\section{Autonomous Reorientation}
 In this section, we introduce the controller used to autonomously reorient the vehicle from an arbitrary orientation on the ground to help it take off. 

\subsection{Problem setup}
\begin{figure}
    \centering
    \includegraphics[width=\linewidth]{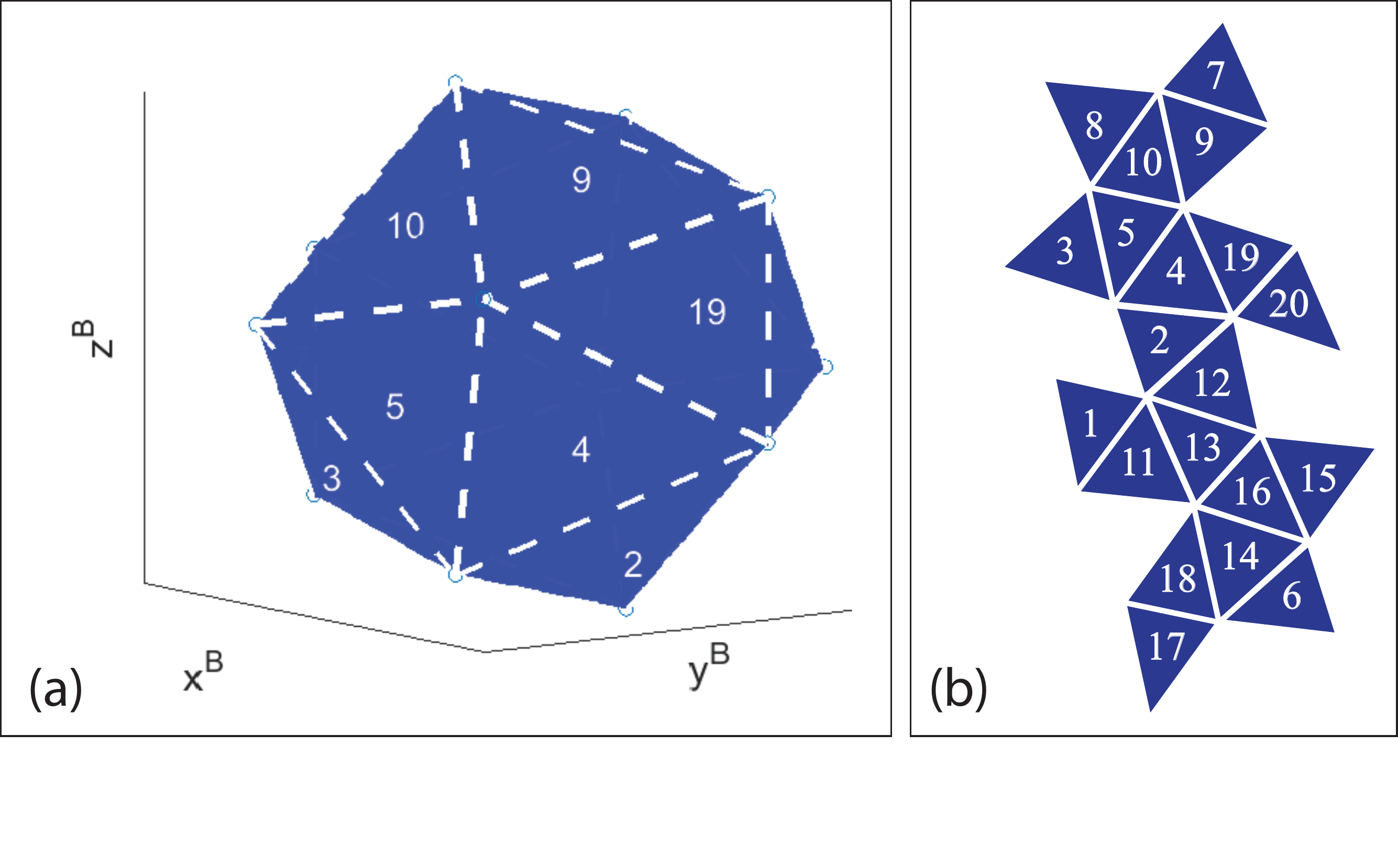}
    \caption{The icosahedron tensegrity structure has 20 faces, numbered from 1 to 20. (a): Isosymmetric view of icosahedron tensegrity. (b): Faces of an unfolded icosahedron tensegrity.}
    \label{TensegrityFace}
\end{figure}

The icosahedron tensegrity has twenty faces, numbered from 1 to 20, as shown in Fig. \ref{TensegrityFace}. After collisions, the tensegrity will land with one of its faces on the ground. Denote the contact face as $F_{i}$ if the $i^{th}$ face of the tensegrity is touching the ground. In most of the attitudes, the propellers are not pointing upwards, and it is very difficult for the tensegrity vehicle to take off. The goal of autonomous reorientation is to find a series of control inputs that can rotate the vehicle back to an attitude from which it can easily take off.

\subsection{Contact face identification}
To successfully rotate to a desired attitude, the vehicle needs to first identify which face of the tensegrity is in contact with the ground. 

The rods of the icosahedron tensegrity structure are parallel with the axes of the body frame. As a result, each contact face of the vehicle corresponds to a specific roll-pitch angle pair that can be found as follows:

Assume the contact face of the tensegrity is $F_i$ and node $j$, node $k$ and node $l$ are the three nodes on the face. Let $\boldsymbol{n}_j^B$, $\boldsymbol{n}_k^B$, $\boldsymbol{n}_l^B$ be the positions of these nodes in the vehicle body frame. Thus, $\boldsymbol{w}_i^B$, a vector normal to the contact face in the vehicle body frame can be found as:
\begin{align}
\boldsymbol{w}_i^B = (\boldsymbol{n}_j^B-\boldsymbol{n}_k^B)\times(\boldsymbol{n}_j^B-\boldsymbol{n}_l^B)
\label{normalU}
\end{align}

Furthermore, a unit vector normal to the contact face and pointing into the icosahedron tensegrity can be computed as: 
\begin{align}
\boldsymbol{v}_i^B = sgn(\boldsymbol{-n}_j^B\cdot \boldsymbol{w}_i^B)\frac{\boldsymbol{w}_i^B}{||\boldsymbol{w}_i^B||}
\label{unitNormalU}
\end{align}
where $sgn$ is the sign function. 

Let $\boldsymbol{z}^B$ be the unit vector pointing along the positive z-axis of the vehicle body frame. The rotational matrix that rotates $\boldsymbol{z}^B$ to $\boldsymbol{v}_i^B$ can be found as follows: 

\begin{align}
\boldsymbol{R}_i = I + S(\boldsymbol{z}^B\times\boldsymbol{v}_i^B) + \big(S(\boldsymbol{z}^B\times\boldsymbol{v}_i^B)\big)^2\frac{1}{1+\boldsymbol{z}^B\cdot\boldsymbol{v}_i^B}
\label{normalV}
\end{align}
Where $S(.)$ is the skew-symmetric cross product matrix, so that $S(\boldsymbol{a})\boldsymbol{b}=\boldsymbol{a}\times\boldsymbol{b}$.

As the attitude of the vehicle is defined as a rotation from the vehicle body frame to the Earth frame, $\boldsymbol{R}_i$ also represents the attitude of a vehicle with contact face $F_i$ and zero yaw angle:
\begin{align}
(0,\theta_i,\phi_i) = f_{\boldsymbol{R}r}(\boldsymbol{R}_i)
\label{findEuler}
\end{align}
where $f_{\boldsymbol{R}r}(.)$ is the conversion from rotation matrix to yaw-pitch-roll Euler-angle set.

Hence, the difference between the estimated attitude and the attitude of a vehicle with contact face $F_i$ and zero yaw angle is:
\begin{align}
\delta\boldsymbol{R}_i = (f_{r\boldsymbol{R}}(0,\theta_i,\phi_i))^T f_{r\boldsymbol{R}}(0,\hat{\theta},\hat{\phi})
\label{findAttitudeDifference}
\end{align}
where $f_{r\boldsymbol{R}}(.)$ converts a yaw-pitch-roll Euler-angle set to a rotation matrix, whereas $\hat{\theta}$ and $\hat{\phi}$ are estimated pitch and roll angles of the vehicle from the attitude estimator introduced in Section III.
Finally, the contact face is identified:
\begin{align}
i = argmin(
f_{\boldsymbol{R}\Theta}(\delta \boldsymbol{R}_i))
\label{identifyContactFace}
\end{align} 
where $f_{\boldsymbol{R}\Theta}(.)$ evaluates the angle of rotation matrix in the sense of the axis-angle representation.

\subsection{Target attitude}
To implement a rotation that switches contact face from $F_i$ to  $F_{i+1}$, the vehicle first has to find its target attitude after the rotation.   

Consider the following scenario: the tensegrity sits on the ground with contact face $F_i$ and the current attitude of the vehicle is estimated to be $\boldsymbol{\hat{R}}$. The desired new contact face is $F_{i+1}$. Notice that face $i$ and face $i+1$ are adjacent. Denote the nodes shared by the two faces as node $j$ and node $k$. Thus, the desired rotation can be represented with an axis-angle pair. The axis, in the Earth frame, can be found as:

\begin{align}
\boldsymbol{e}^E_{i,i+1} = \frac{\boldsymbol{\hat{R}}(\boldsymbol{n}_j^B-\boldsymbol{n}_k^B)}{||\boldsymbol{n}_j^B-\boldsymbol{n}_k^B||}
\label{RotationAxis}
\end{align}
where $\boldsymbol{n}_j^B$ amd $\boldsymbol{n}_k^B$ are positions of node $j$ and node $k$ in the vehicle body frame. Meanwhile, $\Theta$, the angle of rotation between the two adjacent faces can be obtained from geometry as follows: let $\boldsymbol{h}_i$ be a unit vector on face $i$ that is orthogonal to $\boldsymbol{n}_j^B-\boldsymbol{n}_k^B$. Similarly, let $\boldsymbol{h}_{i+1}$ be a unit vector on face $i+1$ that is orthogonal to $\boldsymbol{n}_j^B-\boldsymbol{n}_k^B$. The angle of the rotation can be found as: 
\begin{align}
\Theta = cos^{-1}(\boldsymbol{h}_i\cdot\boldsymbol{h}_{i+1})  
\label{RotationAngle}
\end{align}
Thus, the desired target attitude follows:
\begin{align}
\boldsymbol{R}_d = f_{A\boldsymbol{R}}\big((\boldsymbol{e}^E_{i,i+1},\Theta)\big)\boldsymbol{\hat{R}}  
\label{DesiredAttitude}
\end{align}
where $f_{A\boldsymbol{R}}(.)$ converts an axis-angle pair to a rotation matrix.

\subsection{Reorientation path}
We introduce here a method to find the next contact face during the process of autonomous reorientation.

\begin{figure}
    \centering
    \includegraphics[width=\linewidth]{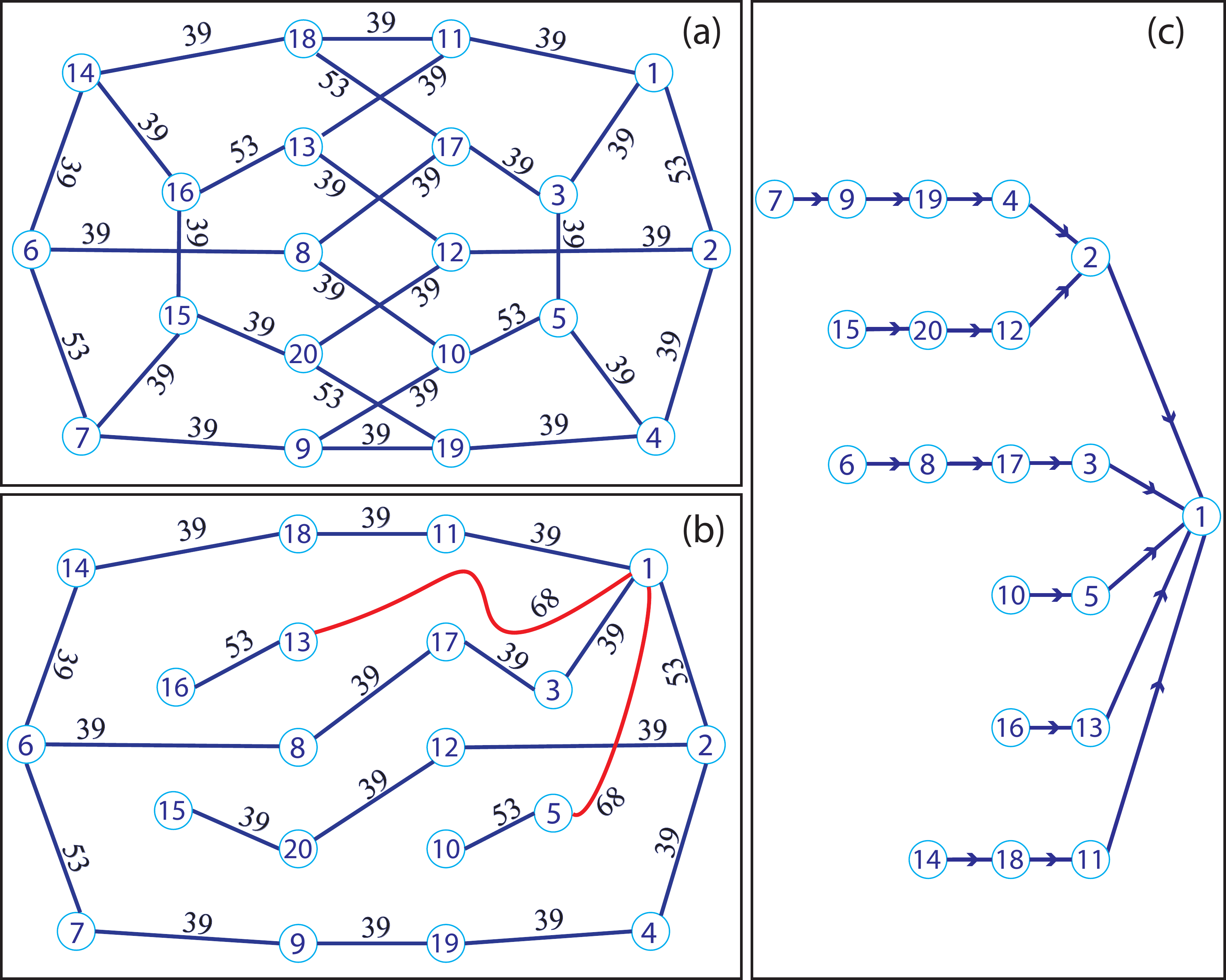}
    \caption{Connectivity plot of the tensegrity vehicle. Circled numbers indicate indices of contact faces. Weights of the edges are angles of the corresponding rotations in degrees. (a): Connectivity plot based on icosahedron geometry. (b): Connectivity plot after deleting infeasible transitions and adding two new rotations about tensegrity nodes (shown as red curves). (c): Final reorientation path.}
    \label{TensegrityConnectivity}
\end{figure}

Possible transitions between contact faces are shown in Fig. \ref{TensegrityConnectivity}. 
Each node in the graph represents a contact face and two points are connected if the vehicle can transition between two faces via rotating about a tensegrity edge. 
The weights of the edges in the graph are the angles between two adjacent faces. We define the contact face $F_1$ as the goal where the vehicle can directly take off. Thus, the objective is to find shortest paths starting from an arbitrary point on the graph to the goal. 

It is worth noting that during the experiment, some paths in the graph are found to be physically infeasible, as they require the vehicle to generate a large yaw torque that exceeds the capability of the vehicle. These connections are deleted from the graph and the operation makes the graph disconnected. To fix the problem, we add two special movements in which the vehicle rotates about a node, instead of an edge, as is represented by the red curve in Fig. \ref{TensegrityConnectivity}.(b).

With the new graph generated, A* search is used to find shortest paths from all possible starting points to the goal. The final reorientation path is illustrated in  Fig. \ref{TensegrityConnectivity}.(c).

\subsection{Implementation of rotations}

Large torques are required to rotate the vehicle on the ground. As a result, the vehicle is modified so that the propellers can be controlled to rotate in both clockwise and counterclockwise directions. Consequently, the propellers can generate both positive and negative thrusts. Thus, the envelop of the feasible body torque generated is broadened. 

During the process of autonomous reorientation, the same attitude controller and thrust converter introduced in Section III is applied. The desired attitude is computed with the method in Section IV.C. Meanwhile, the desired total thrust is set to zero. It is noteworthy that thrust commands are stopped once the vehicle determines its contact face switches from $F_i$ to $F_{i+1}$. At this time, the moment arm of the gravity is expected to be zero-length and the gravity will no longer generate a torque against the rotation. Once the vehicle successfully reaches the desired attitude, it will find its next target with the method discussed in Section IV.C and IV.D. Notice that the controller plans at each step, so it can easily adjust the next target if the vehicle rotates to an unexpected attitude due to disturbance. After finding the next target attitude, the vehicle will perform the planned rotation, and repeat the process until it reaches its goal.

\section{Experimental results}
This section presents the result of the tensegrity design and the experiments demonstrating the vehicle's collision resilience and its ability to autonomously reorient itself. Experiment video: \url{https://youtu.be/9v3vB4RaOew}. 

\subsection{Result of tensegrity design}
We use the analysis tool presented in section II to guide the design of the tensegrity structure. Among all possible candidates satisfying the design requirements, carbon fiber rods with 6mm OD and braided fishing lines are selected based on weight, price and availability. The final tensegrity structure weighs 50g and is about one-fifth of the total 252g vehicle mass. The maximum total thrust of the vehicle is 8.5N, which leads to a 3.4:1 thrust-to-weight ratio.

\subsection{Collision resilience}
We command the vehicle to accelerates towards a wall and to collide with it so as to validate the collision resilience of the vehicle.
\begin{figure}[b]
    \centering
    \includegraphics[width=\linewidth]{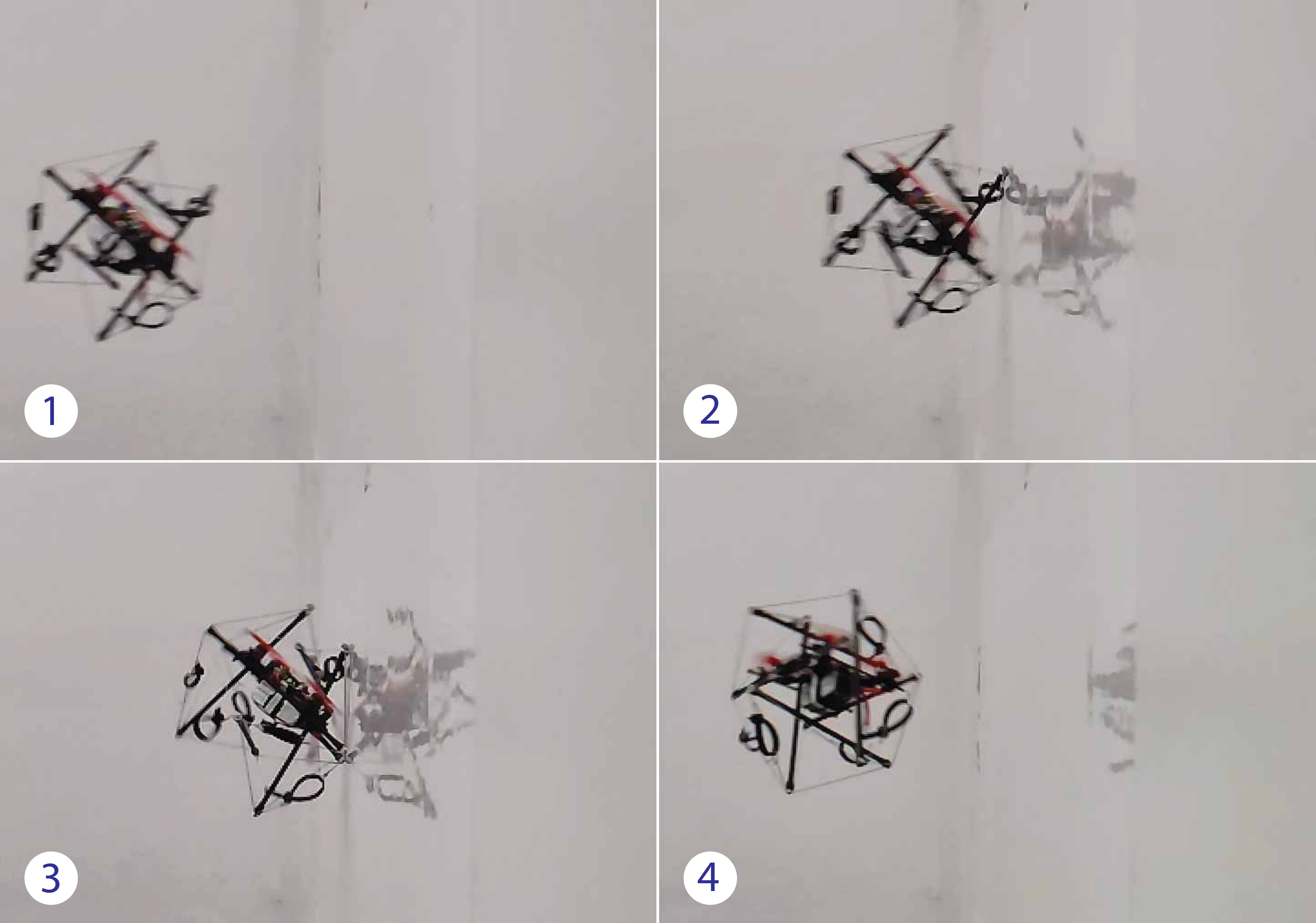}
    \caption{Video sequences of the collision process: 1) Vehicle accelerates towards the wall. 2) Collision starts. 3) Vehicle comes to a full stop. 4) Vehicle bounces back from the wall. The speed of the vehicle right before the collision is 6.5m/s.}
    \label{Collision}
\end{figure}
Video sequences of the collision process is shown in Fig. \ref{Collision}. The vehicle collides with the wall with a speed of 6.5m/s. All parts within the tensegirty frame remain intact throughout the process and the vehicle maintains its ability to fly after the collision.

\subsection{Autonomous Reorientation}
We roll the vehicle like a dice so it starts at a random attitude on the ground. Afterwards, the autonomous reorientation controller is implemented. The vehicle rotates itself to a desired attitude from which it can easily take off.

\begin{figure}
    \centering
    \includegraphics[width=\linewidth]{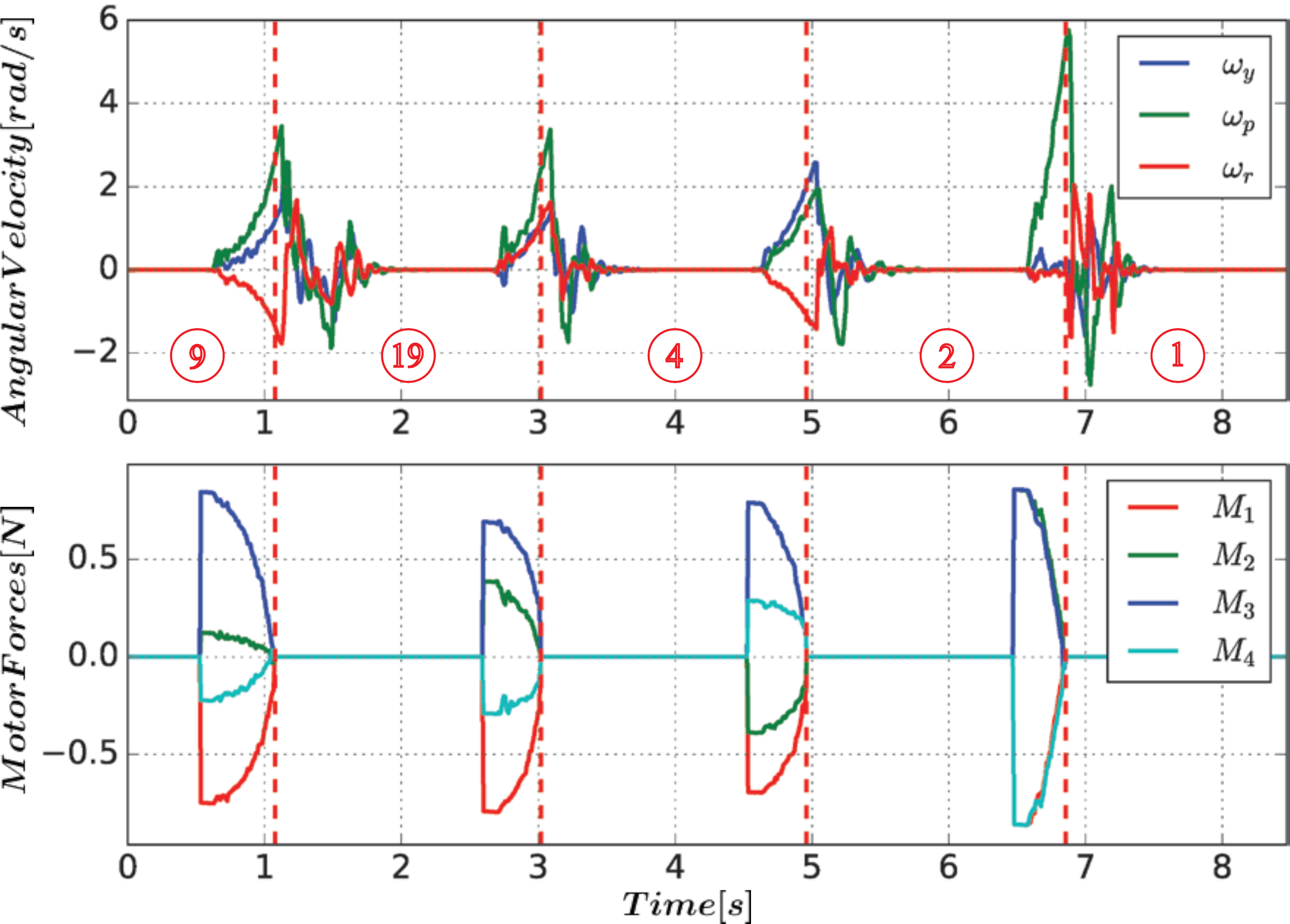}
    \caption{Autonomous reorientation process of the vehicle. The circled numbers are the indices of the identified contact faces whereas red vertical dashed lines indicate the change of contact faces.}
    \label{AutoReorientation}
\end{figure}

Fig. \ref{AutoReorientation} illustrates the details of the autonomous reorientation process. The first row shows the yaw-pitch-roll angular velocity measured by the vehicle's onboard rate gyroscope, and the second row captures the motor thrust commands of the vehicle. Red vertical dashed lines indicate the change of the faces that are identified as the closest to the ground, as described in Section IV.B. The circled numbers are the indices of the identified contact faces. Note that propeller thrust commands become zero once the vehicle determines its contact face has switched. From the graph, it is shown that the vehicle is able to recognize its face on the ground and autonomously rotate itself to the goal attitude.

\section{Conclusion}
In this paper we present the methodology used to design a collision resilient vehicle that adopts the advantages of icosahedron tensegrity structures. We guide the design of the tensegrity with stress analysis of the structure under impact forces during collisions. A cascaded flight controller with a 
state estimator based on an IMU and a motion capture system is proposed to control the vehicle in flight. We also exploit the 20-faced geometry of the icosahedron tensegrity and develop an autonomous controller that divides the reorientation process into a series of rotations switching the tensegrity's face contacting the ground. Thus, the autonomous controller turns a complicated rotation task into a finite state machine that is easy to analyze and implement.

The tensegrity structure resulting from our design methodology weighs about one-fifth of the vehicle's total mass. Due to its light weight, the tensegrity frame causes limited influence on the vehicle's flight performance. The vehicle can still achieve a 3.4:1 thrust-to-weight ratio, which grants decent flight agility.  
Finally, we validate the capability of the vehicle with experiments and show that the vehicle can successfully reorient autonomously and survive collisions with speed up to 6.5m/s.

\section*{Acknowledgements} {The experimental testbed at the HiPeRLab is the result of contributions of many people, a full list of which can be found at \url{hiperlab.berkeley.edu/members/}.
We also kindly acknowledge funds from the UC Berkeley Fire Research Group: \url{https://frg.berkeley.edu/} and the College of Engineering.}


\bibliographystyle{IEEEtran}
\bibliography{references}
\end{document}